\documentclass[conference]{IEEEtran}
\usepackage{times}

\usepackage[numbers]{natbib}
\usepackage{multicol}
\usepackage[bookmarks=true]{hyperref}
\usepackage{graphicx}
\usepackage{booktabs}
\usepackage{amsmath}
\usepackage{xcolor}

\begin{document}

\title{Manipulator-Independent Representations for \\Visual Imitation}

\author{\authorblockN{Yuxiang Zhou}
\authorblockA{DeepMind\\London, UK\\
yuxiangzhou@google.com}
\and
\authorblockN{Yusuf Aytar}
\authorblockA{DeepMind\\London, UK\\
yusufaytar@google.com}
\and
\authorblockN{Konstantinos Bousmalis}
\authorblockA{DeepMind\\London, UK\\
konstantinos@google.com}}

\maketitle

\begin{abstract}
Imitation learning is an effective tool for robotic  learning tasks where specifying a reinforcement learning (RL) reward is not feasible or where the exploration problem is particularly difficult. Imitation, typically behavior cloning or inverse RL, derive a policy from a collection of first-person action-state trajectories. This is contrary to how humans and other animals imitate: we observe a behavior, even from other species, understand its perceived effect on the state of the environment, and figure out what actions our body can perform to reach a similar outcome. In this work, we explore the possibility of third-person visual imitation of manipulation trajectories, only from vision and without access to actions, demonstrated by embodiments different to the ones of our imitating agent. Specifically, we investigate what would be an appropriate representation method with which an RL agent can visually track trajectories of complex manipulation behavior ---non-planar with multiple-object interactions--- demonstrated by experts with different embodiments. We present a way to train manipulator-independent representations (MIR) that primarily focus on the change in the environment and have all the characteristics that make them suitable for cross-embodiment visual imitation with RL: cross-domain alignment, temporal smoothness, and being actionable. We show that with our proposed method our agents are able to imitate, with complex robot control, trajectories from a variety of embodiments and with significant visual and dynamics differences, e.g.\ simulation-to-reality gap. 
\end{abstract}

\IEEEpeerreviewmaketitle
\section{Introduction}
Primates and especially humans depend on understanding actions of others to survive, socially organize, and, in the case of humans, learn new skills by imitation~\cite{rizzolatti2004mirror}. Thanks to our visuomotor ``mirror'' neurons, we are able to observe behavior by others, process that visual information, and map it to our own embodiment in order to perform similar actions that arrive at the same results. Replicating a similar imitation performance in robots will surely increase their usefulness, interaction capacity with humans, and significantly decrease the cost of learning new skills.
We are motivated by the prospect of a robotic manipulator that can imitate any visually-demonstrated behavior (a sequence of goals) of arbitrary complexity. These goals will need to be perceived, and to provide a context with which an imitating agent can act towards reaching them. 

\begin{figure}[t]
\centering
\includegraphics[width=\linewidth]{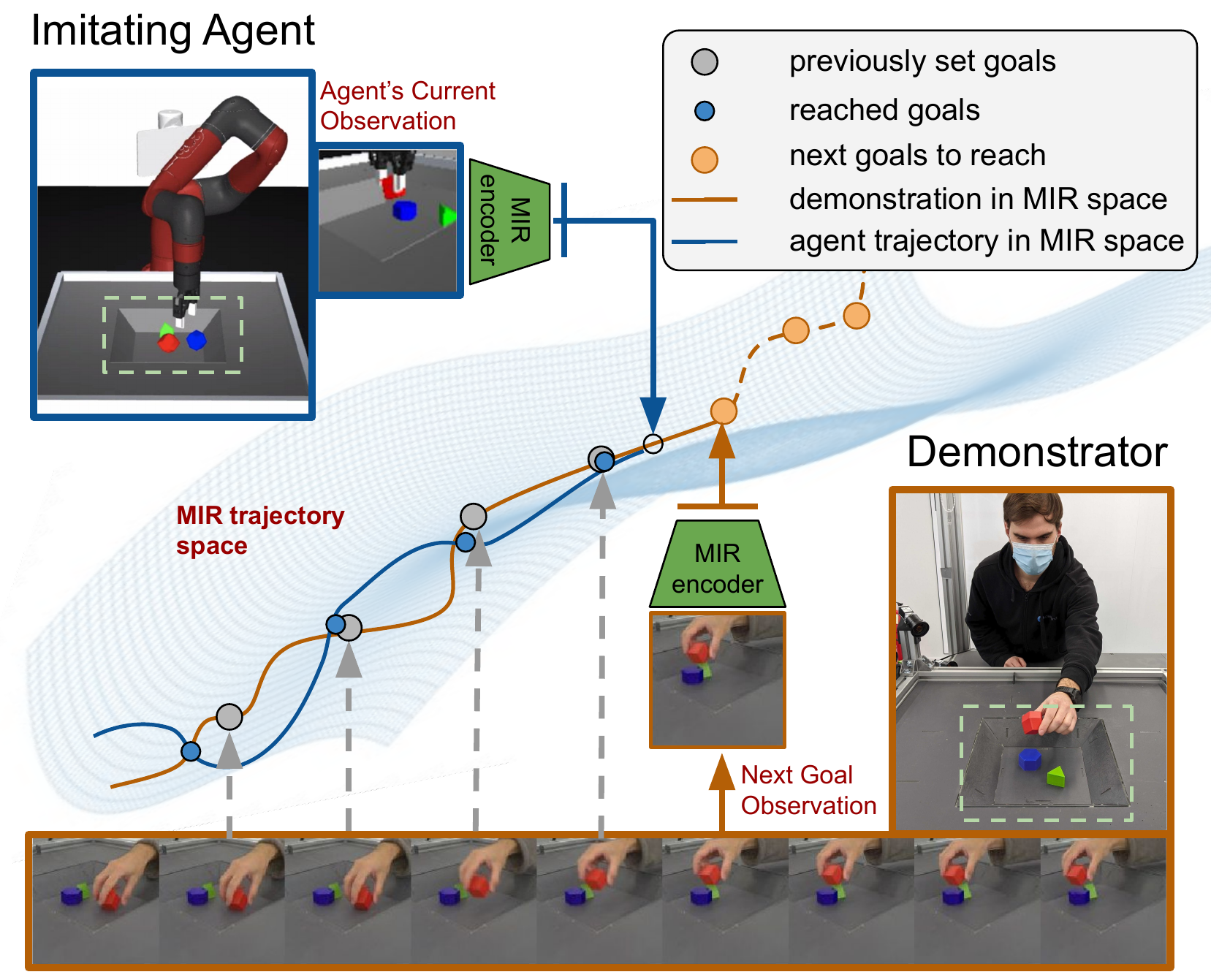}
\caption{Our work focuses on learning manipulator-independent representations (MIR) that can be used for imitating trajectories of behavior demonstrated with different embodiments, unseen during training, solely from pixel observations, even in the presence of large visual domain gaps.
}
\vspace{-0.6cm}
\label{fig:teaser}
\end{figure}

In this work we focus  on imitating single trajectories as precisely as possible, \textit{using only vision}, without access to proprioceptive information or actions of the demonstrator\footnote{For video examples please visit \url{https://sites.google.com/view/mir4vi}}. Our primary goal is to explore specifically how we can provide robots with the \textit{perceptual} ability to mirror a demonstrated behavior as accurately as possible, even when that is performed by a ``demonstrator'' with a different embodiment or in a different domain (simulation vs.\ reality). In other words we want to equip our robots with the ability of third-person imitation, particularly specialized for cross-embodiment scenarios.  We tackle this from a manipulation viewpoint: we learn  representations  that capture the change effected in the world and details related to it, but ignore some manipulator-specific information, like the joint angles of a specific robotic arm. A characteristic example from our experiments, shown in Figure~\ref{fig:teaser}, is imitating trajectories demonstrated by a human hand, an environment and embodiment both unseen during our perceptual training.

Although the perceptual methods we propose are agnostic to the specific formulation of the cross-embodiment visual imitation task, we study the problem through trajectory tracking with reinforcement learning (RL)~\cite{peng2018deepmimic, aytar2018playing, peng2018sfv}. General solutions for reaching visual goals, like goal-conditioned policies~\cite{pathak2018zero}, have been shown to work only for simple, short-horizon trajectories in manipulation, and with simple control for the imitating agent~\cite{pathak2018zero, sharma2019thirdperson}. If the behavior demonstrated involves multiple objects with rich dynamics or is not (almost) planar, goal-conditioned visual imitation is much more difficult and has not been shown to work, to the best of our knowledge. The same applies for imitating agents with complex control, for example with actions that correspond to joint velocities, as in this work, as opposed to end-effector positions.
For that reason, prior visual imitation work~\cite{Liu2017imitation} has leaned on trajectory tracking: learning the necessary sequences of actions to regenerate a demonstrated trajectory. %
This high-fidelity imitation task is in contrast to cloning policies of a demonstrator with different observations but known actions~\cite{stadie2017third}. It is also different from one-shot visual imitation, which focuses on generalizing or adapting existing knowledge to a slight variation of previously seen tasks~\cite{Liu2017imitation,Duan2017oneshot,dasari2020transformers, finn2017one}. 
This then begs the question: how do we learn pixel-based representations that produce the right information for a cross-embodiment trajectory tracking agent? In this work we identified three  characteristics: \textsl{(a)} cross-domain alignment, \textsl{(b)} temporal smoothness,  and \textsl{(c)} actionable representations. 

Our solution is to learn a common embedding space which has the capacity of aligning trajectories with large domain gaps. We propose training manipulator-independent representations (MIR) in a cross-domain way with a combination of a self-supervised time-contrastive loss, for temporal smoothness, and a goal-conditional skill learning objective, for actionable representations. In order to obtain aligned sequences for cross-domain training, but also to achieve better transferability to the real world and unseen manipulators, we utilize several levels of domain randomization in our simulated environments. This includes an `invisible arm' environment where the objects move but the manipulator is invisible, as shown in Figure~\ref{fig:ctg_domains}. This helps the representation to focus on the actual change in the environment, while other types of randomization help capture the rough position and properties of a manipulator. 

Our main contribution, in this work, is a method and a training regime with which we can learn manipulator-independent representations (MIR) for trajectory tracking.  Prior work~\cite{james2019sim, sharma2019thirdperson} dealing with cross-embodiment visual imitation has been assuming use of data from the demonstrator's environment during perceptual training, and has been focusing on one-shot imitation and not high-fidelity trajectory imitation. Our  method for learning MIR representations is a general one, i.e. it is not utilizing specific task or demonstrator information. We show that our MIR approach outperforms other reasonable choices and is able to deal with large domain gaps, difficult control, and complex object dynamics.  This is also the first work, to the best of our knowledge, on cross-embodiment trajectory tracking, i.e.\ visual imitation of robotic manipulation trajectories from unknown embodiments. 

The paper is structured as follows: we first provide some background on third-person visual imitation in Section~\ref{sec:background}, particularly in the context of robotics manipulation from pixels. We then describe, in Section~\ref{sec:CEVI}, cross-embodiment visual imitation and the details of our RL based trajectory tracking method. In Section~\ref{sec:data} we discuss the environments, generated dataset, and demonstrated trajectories. In Section~\ref{sec:results} we give an overview of the evaluations of our claims and discuss our results, and finally we conclude with Section~\ref{sec:conclusion}.

\section{Related Work}

\label{sec:background}

Imitation learning is a problem with a wide range of methodologies in literature depending on the precise problem formulation and assumptions. Based on the amount and the kind of demonstration data assumed to be available, we can roughly split existing methods into two categories. There are those that assume ample demonstration data, with known actions, and attempt to generalize a policy from them. Then there are those that use a single previously unseen demonstration, usually with unknown actions, as the specific behavior the agent is asked to imitate.

In the first category, there is an implicit assumption of an intent or task that the imitating agent needs to learn. Two main streams of such works are behavioral cloning~\cite{bain1995framework, ross2011reduction} and inverse reinforcement learning~\cite{russell1998learning, ng2000algorithms, finn2016guided, ho2016generative}. The former deals with imitating an expert, typically defined by a set of demonstrations with known actions and observations similar to the ones observed by the imitating agent. Inverse RL deals with the problem of recovering a reward function from demonstrated data and training an RL agent with such a reward. The behavior is defined and restricted by the type of demonstration data collected. 

In the second category the agent is provided with a trajectory that defines the imitation task. Here we can include works in one-shot imitation and trajectory tracking, which is where our work also belongs. One-shot imitation deals with generalizing or adapting to a novel task at test time. This is typically defined by a first-person trajectory of a slight variation of previously seen tasks, e.g.\ a new viewpoint or a new scene configuration~\cite{Liu2017imitation,OpenAI2021asymmetric, Duan2017oneshot,dasari2020transformers, finn2017one}. 
While one-shot imitation is about generalizing across task variations, trajectory tracking is about high-fidelity replication of a specific sequence of goals. Feature tracking of trajectories~\cite{peng2018deepmimic, aytar2018playing, peng2018sfv, paine2018one} with RL has been successfully used primarily for locomotion and with hand-engineered features. Tracking from such features is not always feasible and is mostly based on task-specific solutions (e.g.\ mocap data). 

Using vision for tracking is desirable but expensive if doing it end-to-end from pixels with RL, even when the demonstrator and imitator domains are known~\cite{pathak2018zero, Liu2017imitation}. Visual trajectory tracking for manipulation tasks introduces additional challenges due to the stochasticity of the interaction with the unactuated objects. 
In our work we also use RL for visual tracking of manipulation trajectories demonstrated with an embodiment and domain unknown during our perceptual training. Although a generalized trajectory-conditioned imitation solution akin to goal-conditional policies would be desirable, they have been shown to be effective only in single-embodiment cases, for simple near-planar tasks~\cite{pathak2018zero} and primarily from features~\cite{lynch2020learning, OpenAI2021asymmetric}.

Our biggest challenge is dealing with the domain gap in the manipulator (e.g.\ human hand vs. robotic gripper), in the low-level characteristics of the visual observations, and even in the physics (e.g.\ simulation vs.\ reality). Current imitation learning approaches from pixels are mainly tested within a single domain~\cite{pathak2018zero,peng2018deepmimic,mandlekar2020learning,yu2018one}, domains with a limited gap~\cite{aytar2018playing, Liu2017imitation, stadie2017third}, or require manually labelled data~\cite{peng2018sfv}. For example, the third-person imitation work by ~\citet{stadie2017third} deals with known actions, observations from the same simulated embodiment, but with a different viewpoint. On the other hand, imitation learning for large domain gap happens in model dependent state space and primarily for locomotion tasks~\cite{merel2018neural, hasenclever2020comic, peng2018sfv,peng2020learning,peng2018deepmimic}.

The works closest to ours are \citet{Liu2017imitation, sharma2019thirdperson, smith2020avid} and \citet{james2019sim}. Similar to our work \citet{Liu2017imitation} deals with RL-based trajectory visual tracking with a domain gap. In this particular case the gap has to do with a viewpoint difference. Although the demonstrations are performed by a human, the observations are carefully cropped to leave the manipulators out of the scene. Both  \cite{sharma2019thirdperson, smith2020avid} deals with cross-embodiment demonstrations in the context of one-shot imitation. \citet{sharma2019thirdperson} uses temporally aligned human and robot demonstration videos in order to learn how to propose goals for robot given a human video. They operate in a predefined task space and scaling this method to general imitation would be quite challenging as it requires manually collected aligned trajectories. In \cite{smith2020avid} a technique using CycleGAN~\cite{zhu2017unpaired} is proposed to translate human demonstrations to robot-looking ones at pixel space.
More similar to our work, in \citet{james2019sim}, domain randomization and models of human hands are used in simulation to be able to train the perceptual stack to generalize to human hand demonstrations in the real world. An important difference to all of these works is that we do not assume a priori the knowledge of the demonstrator domain. With the exception of \cite{james2019sim}, these works use data collected in the demonstrator domain as part of their perceptual training process. \citet{james2019sim} nevertheless assumes a priori that the demonstrations will be from a human hand. In contrast, our representation are manipulator-independent and our proposed method can successfully track demonstrations from a variety of previously unseen manipulators.  

\begin{figure}[t]
\centering
\includegraphics[width=0.5\textwidth]{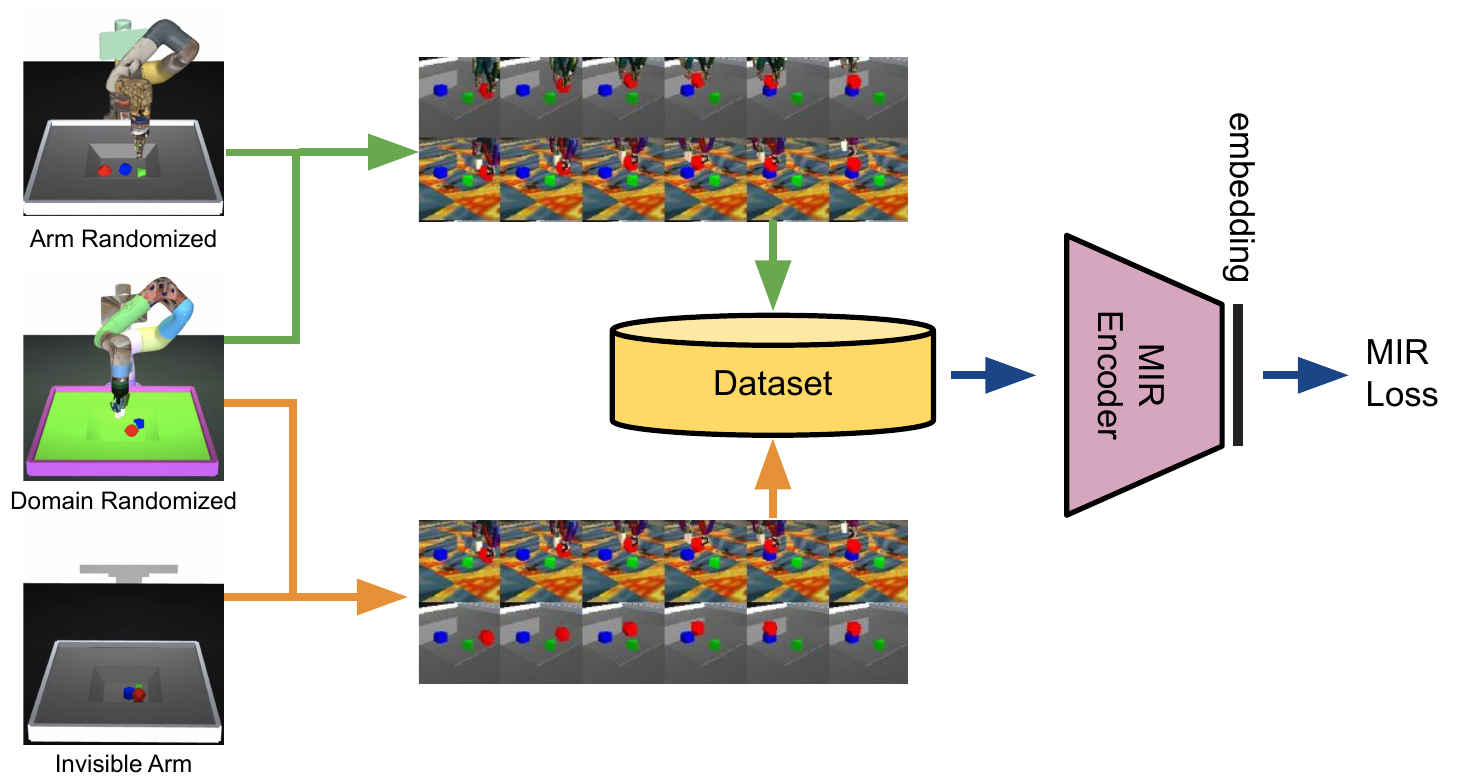}
\caption{Learning manipulator-independent representation (MIR) space. MIR is trained on a dataset generated using two pairs of environments: (a) Domain-randomized and `invisible arm' environment and (b) domain-randomized and arm-only-randomized environments. Please see Section \ref{sec:dataset} for the details of dataset collection.
}
\vspace{-0.6cm}
\label{fig:methods}
\end{figure}

\section{Cross-embodiment visual imitation}
\label{sec:CEVI}
In this section we explain how we learn to imitate unconstrained manipulation trajectories executed by previously unknown manipulators, such as a human hand,
using only \emph{visual observations}. 
Our imitation method is composed of two main phases: \textsl{(a)} learning a manipulator-independent representation (MIR) space (see Figure~\ref{fig:methods}), and \textsl{(b)} cross-embodiment visual imitation through reinforcement learning using the pre-trained MIR space (see Figure~\ref{fig:teaser}).

\subsection{Manipulator-Independent Representations}
\label{sec:MIR}

Perhaps the most crucial part in visual imitation by trajectory tracking is to have a high-level feature space that can match demonstrator and agent observations adequately. In this work we address learning such a space that specializes in manipulator-independent imitation. We identified three main properties of the desired feature space that are crucial for successful imitation from unseen manipulator trajectories. These are \textsl{(a)} cross-domain alignment, \textsl{(b)} temporal smoothness, and \textsl{(c)} being actionable (suitability for RL). 

{\bf Cross-Domain Alignment}.
One of the most daunting challenges in cross-embodiment imitation is the \emph{domain gap} between the demonstrator trajectory and the agent's own environment.
The domain gap could be due to the change in observations (e.g. viewpoint, simulation vs.\ reality), the manipulator (e.g. agent morphology and action space) or dynamics of the environment (e.g. simulation vs.\ reality). We deal with the domain gap by learning manipulator-independent representations with the help of domain-randomized simulated environments and domain alignment methods.

The main target in domain randomization \cite{sadeghi2016cad2rl,tobin2017domain} is to randomize the observations and dynamics of the simulated environment with the hope that an agent trained under this setting will be robust to real-world visual observations and dynamics. In domain randomization the representation space and the policy are typically tightly coupled. Therefore the manipulator's body, action space and the task at hand are strongly ingrained in the representation space. In our work we decouple the domain-randomized representation from manipulator type and task-specific features. Hence the learned representation space can be utilized for any task with any manipulator. This is a crucial step in our work as the target is to transfer behavior across manipulators with unobserved (e.g. human hand) morphology and action space without being constrained by any task. In order to achieve that, we need a representation space that is both domain-invariant and can capture a good high-level understanding of the environment. 

\begin{figure*}[t]
\centering
\includegraphics[width=\textwidth]{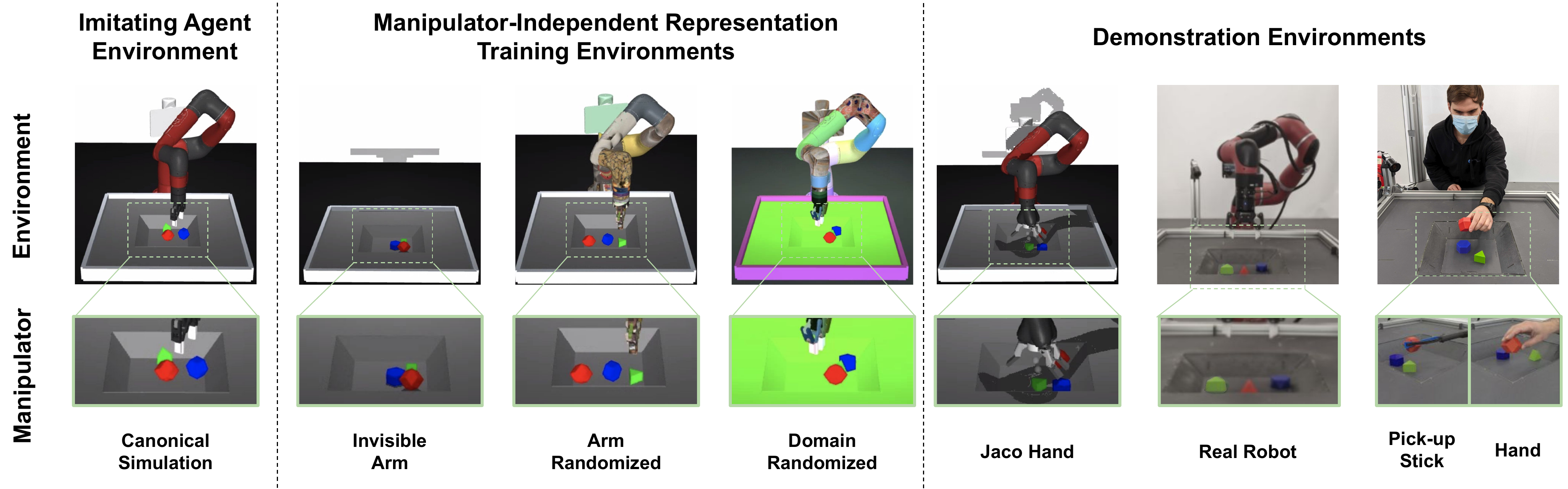}
\caption{
\textbf{Visualization of the different environment used in our work.} Our imitating agent always operates in a canonical simulated environment (left). We have 3 additional versions of it which are used to generate data for perceptual training (middle). Finally we use 4 held-out demonstration domains, both in simulation and in the real world, for cross-embodiment imitation.}
\label{fig:ctg_domains}
\vspace{-0.6cm}
\end{figure*}

In order to learn such an abstract representation for cross-embodiment imitation, we need to consider what that entails. When a robot imitates a trajectory, it can either \textsl{(a)} mimic the movements of the manipulator, or \textsl{(b)} replicate the effect of the manipulator on the environment. To illustrate this dilemma consider the demonstrated trajectory of stacking one object onto another. Assume that during imitation the robot has grasped the first object but dropped it halfway while moving to the other object. Now the robot has to make a decision between following the path of the imitated manipulator blindly and picking up the dropped object. The latter would mean it prioritizes matching the changes in the environment more than following the manipulator. While learning MIR we make sure the representation can capture the changes in the environment without ignoring the rough motion of the manipulator. We achieve this by training our domain alignment methods on two pairs of environments: \textsl{(a)} the `domain-randomized' and `invisible arm' environments which help better capture the changes in the environment; and \textsl{(b)} the `domain-randomized' and `arm-randomized' environments which further make sure that our representations encode information about the manipulator without paying attention to its specific characteristics. These environments are illustrated in Figures~\ref{fig:methods} and~\ref{fig:ctg_domains}, and discussed in more detail in the Section~\ref{sec:results}.

{\bf Temporal Smoothness} implies that observations that are temporally neighbors have similar representations, even when they are from different domains. It is an important property particularly if the representation space will be used for reaching goals specified through visual observations. Using pairs of temporally-synchronized trajectories from the two domain pairs mentioned above we learn our temporally-smooth representations, which we call Temporally-Smooth Contrastive Networks (TSCN), by building upon the widely-used Time-Contrastive Networks (TCN)~\cite{sermanet2018time}. 
\begin{figure}[h]
\centering
\includegraphics[width=0.5\textwidth]{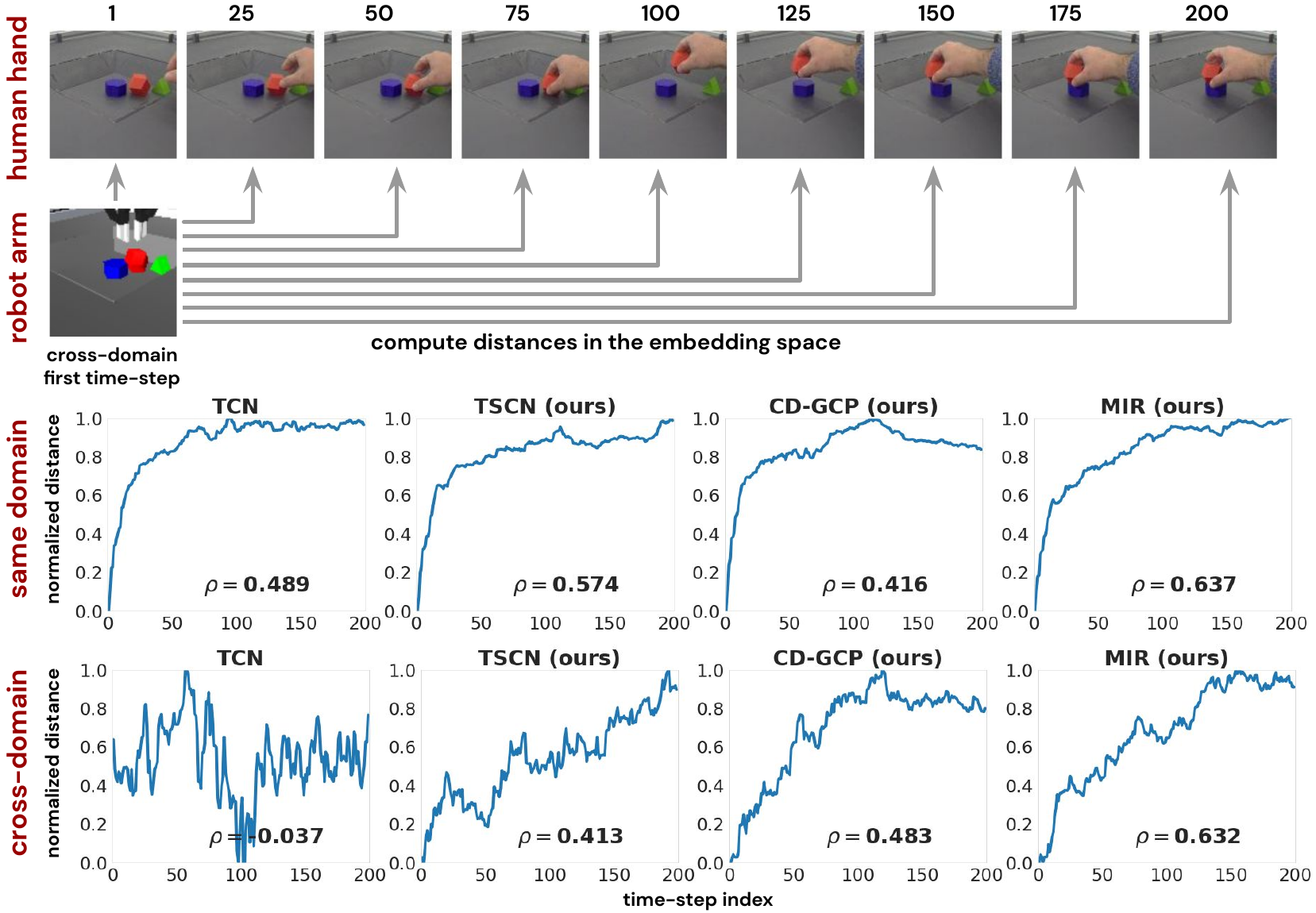}
\caption{{\bf Embedding space distances for same-domain and cross-domain goals.} The first time-step is assumed to be the current observation and the goals are selected over the entire demonstration sequence with increasing temporal distance. Note that objects are manually aligned in simulation to match the first observation in the real-world sequence. The distance plots are normalized to $[0,1]$ for each trajectory and aggregated across $10$ trajectories. We also provide the Spearman's rank correlation between the reachability distance (i.e. linear increase of distance over time) and embedding distance for each of the trajectories. The mean rank correlations over all $10$ trajectories are displayed on each plot. Note that TCN performs reasonably well in the same domain but not across domains. MIR and its two components separately (i.e. TSCN and CD-GCP) are significantly better correlated with the reachability across domains. MIR has similar performance within and across domains.}
\label{fig:reachability}
\vspace{-0.3cm}
\end{figure}

TCN learns representations by enforcing higher similarity between temporally-aligned observation pairs across two different camera views, compared to any other pair selected within a single-view sequence. In our context the views are the two randomized domains that we want to align.  Formally, given two temporally-aligned trajectories of length $N$ from two domains $\mathbf{o}=\{o_i\}_{i=1}^N$ and $\mathbf{\bar{o}}=\{\bar{o}_i\}_{i=1}^N$, we first apply the encoder $\phi$ to obtain the latent representations $x_i = \phi(o_i)$ and $\bar{x}_i = \phi(\bar{o}_i)$ for any observation. Given the aligned encoded sequences $\mathbf{x}$ and $\mathbf{\bar{x}}$, the TCN objective that we minimize is
$\min_{\phi} \left(-\sum_i^N \log \, \frac{ \exp(x_i^{T} \bar{x}_i)} {\sum_{j}^N \exp(x_i^{T} \bar{x}_j)}\right)$.
We particularly use an $n$-pairs implementation as opposed to the triplet implementation due to its robustness \citep{sermanet2018time}. In this work we chose $n=50$. $n$-pairs TCN essentially classifies the matching pair among all other non-matching negative pairs within the given two aligned trajectories. 
One drawback of TCN is that it penalizes misclassification of the negative pairs $(x_i,\bar{x}_{i+1})$ and $(x_i,\bar{x}_{i+50})$ equally, though it is more acceptable to confuse temporally nearby observations as opposed to temporally distant pairs.

In order to incorporate this observation in the loss function of TSCN, we update the softmax cross-entropy objective by providing a temporally-smooth distribution $p_i$ as shown below:
\begin{equation}
 \min_{\phi} \left ( -\sum_i^N \sum_k^N p_{ik} \log \, \frac{\exp(x_i^{T} \bar{x}_k)}{\sum_{j}^N \exp(x_i^{T} \bar{x}_j)}\right), \label{tcn_npairs2} \\ 
\end{equation}
\begin{equation}
 \quad \text{where} \quad  p_{ik}=\frac{exp(-|i-k|)}{\sum_u^N exp(-|i-u|)}  \nonumber
\end{equation}
This objective encourages the learned representation to be more temporally smooth. We also include negative pairs that are populated from other sequences in the same mini-batch which is commonly applied in many pair-based contrastive learning objectives \cite{chen2020simple, tian2019contrastive}. This further improves the alignment quality in the presence of distractors from other episodes.

Moving one step further from temporal smoothness, it is desirable if the distance between the current and goal observations in the embedding space correlates well with the \emph{reachability distance}, the minimum time required to move from the current state to the goal state. Of course the reachability distance is very hard to measure as we don't have optimal policies that can take us from one state to another in the shortest available time in the real world. However, assuming that humans are near optimal agents, we can utilise human trajectories to analyse the relationship between the embedding distance and reachability distance. We collected $10$ stacking trajectories with different object configurations where a human hand reaches one colored object and stacks it on top of another colored object. Considering there is no going back and forth in a stacking trajectory, it is safe to assume that reachability distance between the first time-step and all the other time-steps are linearly increasing over time for human hand trajectories. In Figure~\ref{fig:reachability} we illustrate the embedding space distance between the first time-step and all the remaining time-steps both within and across domains. Note that objects are manually aligned in simulation to match the first observation in the real-world sequence. A linear increase of distance would mean a great correlation between the embedding space distance and the reachability distance. As it is clear in the Figure~\ref{fig:reachability}, cross-domain distances in our TSCN embedding space has a much smoother increase over time as opposed to the TCN method. For a quantitative measurement we also computed the spearman's rank correlation between the reachability distance (i.e. linear increase of distance over time) and embedding distance for each of the trajectories, and the mean rank correlation over all $10$ trajectories are printed on the plots for each method.

{\bf Actionable Representations.}
Another desired property of a good representation space is being actionable, i.e. amenable for RL training. This requires encoding actions while learning the representation. Note that both TCN and TSCN do not utilise actions. To accommodate this we used offline goal-conditioned policy trained in a cross-domain way as an additional method to enrich our representation space. As we have cross-domain aligned trajectories in our dataset, we also introduce offline cross-domain goal conditioned policy training as a way to increase robustness across the domains. Goal conditioned policies are trained to reach a goal that is provided as part of the input \cite{kaelbling1993learning, nasiriany2019planning, pathak2018zero}. While in theory they can reach any goal, in practice their effectiveness is constrained to nearby goals~\cite{nasiriany2019planning}.

Given the current observation $o$ and a goal observation $g$ that is reachable in the near future, a goal conditioned policy $\pi(o, g)$ predicts one step of action $\hat{a}$ that would bring the agent closer to $g$. We can train such a policy using a large set of episodes collected from the environment. Formally, given any $N$-step sequence $\{(o_i, a_i), (o_{i+1}, a_{i+1}), ... , (o_{i+N}, a_{i+N})\}$ extracted from an episode, we take $o_i$ as the current observation and pick a goal randomly from one of the $N$ observations in the future. Then we simply train the goal conditioned policy by minimizing the squared Euclidean loss $||\pi(o_i, o_{i+j}) - a_i||^2$ where $1 \leq j \leq N$.

The extension to our Cross-Domain Goal-Conditional Policies (CD-GCP) is then straightforward. Given a paired $N$-step sequence $\{(o_{i+j}, a_{i+j})\}_{j=0}^N$ and $\{(\bar{o}_{i+j}, \bar{a}_{i+j})\}_{j=0}^N$ from two domains, we train a cross-domain goal conditioned policy $\pi_\times$ by minimizing the loss $||\pi_\times(o_i, \bar{o}_{i+j}) - a_i||^2$ where $1 \leq j \leq N$. Note that the goals in this objective come from the other domain. We set $N=20$ in our experiments. 

In order to obtain our final \textbf{MIR} space we train CD-GCPs together with TSCN. The final CD-GCP loss is $||\pi_\times(\phi(o_i), \phi(\bar{o}_{i+j})) - a_i||^2$ where the embedding network $\phi$ is shared with the TSCN objective described in Equation \ref{tcn_npairs2}. As a result the learned representation becomes more effective as the observation and goal embeddings share the same space, which is optimized for reducing the domain gap. All the objectives introduced so far are optimized for all of the aligned sequences in our dataset, which is described in Section \ref{sec:dataset}. 
As CD-GCPs are trained for reaching cross-domain goals, the embedding distances in the learned representation space also correlates well with reachability distance. Moreover, the MIR space, which a combination of both TSCN and CD-GCP, achieves the best rank correlation for reachability distance prediction between the simulated environment observation and real-world human hand trajectories as shown in Figure~\ref{fig:reachability}. 

\subsection{Cross-Embodiment Imitation via RL}
\label{sec:imitation}

In this section we describe how an RL agent operating in a simulated environment learns to follow a given trajectory utilizing the frozen pre-trained MIR embedding space. Given a cross-embodiment trajectory  $\{\bar{o}_i\}_{i=1}^N$ as a sequence of observations to be followed, we first uniformly sample goals $\{\bar{g}_i\}_{i=1}^M$ where $M < N$, and obtain their embeddings by running them through the encoder $\phi(\bar{g}_i)$. In practice, we observed that the optimal goal sampling rate is every 5 to 10 observations. Only one goal is visible to the agents at anytime. Once an agent reaches the goal, the next goal becomes the active one, as illustrated in Figure~\ref{fig:teaser}. The agent should reach each intermediate goal sequentially till the end to claim a successful attempt. We define our goal-based reward as described below:

\[
    r(o, \bar{g}) = 
\begin{cases}
    1,  & \text{if } \exp(-w||\phi(o)-\phi(\bar{g})||^2) > \epsilon\\
    0,  & \text{otherwise}
\end{cases}
\]
where $o$ is the current observation of the agent and $\bar{g}$ is the next available goal. $w$ serves as the normalization factor which is defined as the mean of the element-wise Euclidean distances between the neighboring encoded observations in the demonstration. $\epsilon$ is randomly sampled in range [0, 1] during parameter sweep and turned out to be optimal around 0.3. Once the agent meets the current goal (i.e.\ reward above $\epsilon$), the next goal becomes visible to the agent. In this work, we use Maximum a Posteriori Policy Optimization (MPO)~\cite{abdolmaleki2018maximum} for training our RL agent.

\section{Environments and Data}
\label{sec:data}
In this section, we present all the environments used for our experiments. We also discuss details about the dataset used for manipulator-independent representation (MIR) training. Finally, we describe the trajectories used to evaluate MIR and other baselines, when used for RL imitation.
\subsection{Environment Details}
Our experiments involve 8 domains, as illustrated in Figure~\ref{fig:ctg_domains}. Our imitating agents always act in a \textbf{canonical} MuJoCo~\cite{todorov2012mujoco} simulated environment. This is designed to have similar appearance and dynamics to our equivalent real environment with a Sawyer arm and a Robotiq 2-finger gripper. The objects used are three geometric props generated procedurally. Two cameras are placed at the front left and right of the basket to generate the $128\times 128$ pixels observations the agents use. \\

As a slight variation of our canonical simulation environment the Sawyer and the Robotiq are rendered \textsl{visually} \textbf{invisible}. This environment is  used for  learning our manipulator-independent representations that focus on environmental changes and ignore the manipulator.
In order to aid MIR training we also use two randomized versions of our canonical environment. In both versions we randomize colour, texture, lighting, and camera poses to create a large visual diversity. Physics properties (e.g. mass, friction, armature, damping, gear) are perturbed uniformly within $\pm10\%$ of their respective values in the canonical environment. In the first, \textbf{arm-randomized} version, we only apply the randomizations on the arm and the gripper. In the second, general \textbf{domain-randomized} environment, those are applied on every entity in the simulated environment.

\begin{table*}[h]
\centering
    \begin{tabular}{l||r|r|r|r|r||r|r|r|r|r}
    \toprule
    & \multicolumn{5}{c||}{\textbf{Lifting Success}}
    & \multicolumn{5}{c}{\textbf{Stacking Success}}  \\
        &  \textbf{invisible} &   \textbf{jaco} &\textbf{robot} & \textbf{stick}& \textbf{hand}
        &  \textbf{invisible} &   \textbf{jaco} &\textbf{robot} & \textbf{stick}& \textbf{hand}  \\
    \textbf{Method}       &        &        &       &       &       &       &       &       &      \\
    \midrule
    \quad TDC \cite{aytar2018playing}           
                    &   67\%            &  38\%           & 10\%                     & 21\%                         & 31\%
                    &    0\%            &   2\%           & 0\%                     & 6\%                         & 0\% \\
    \quad GCP       
                    &   \textbf{100\%}  &  50\%           & 39\%                  & 13\%                      & 10\%  
                    &   0\%             &  0\%            & 0\%                   & 0\%                       & 0\%  \\
    \midrule
    \quad TCN \cite{sermanet2018time}         
                    &   79\%            &  85\%           & 27\%                  & 12\%                       & 0\%  
                    &   0\%             &  0\%            & 0\%                   & 0\%                       & 0\%  \\
    \quad CMC \cite{aytar2018playing}         
                    &   80\%            &  \textbf{100\%} & 80\%                  & 26\%                      & 0\%  
                    &   20\%            &  66\%           & 0\%                   & 0\%                       & 0\%  \\                    
    \quad MIR (ours) 
                    & \textbf{100\%}    &  \textbf{100\%} & \textbf{100\%}        & \textbf{44\%}             & \textbf{50\%}   
                    & \textbf{38\%}     &  \textbf{81\%}  & \textbf{29\%}         & \textbf{17\%}             & \textbf{11\%} \\
    \bottomrule
    \end{tabular}
\caption{Quantitative comparison of representation methods for cross-embodiment imitation. Evaluation is performed on 100 attempts of following for each of the 10 demonstrated stacking trajectories on five domains: invisible-arm, jaco hand, real robot, pick-up stick and human hand. The metrics are the success rate of imitating lifting and stacking behaviours.}
\label{tab:mir_comparison}
\vspace{-1cm}
\end{table*}

For our cross-embodiment imitation experiments we use 4 domains unseen during MIR training. These are presented in order of increased difficulty, in terms of the differences to the three environments above. Firstly, we evaluate our method with trajectories from a version of our canonical simulated environment which has the 2-finger Robotiq replaced by a 3-finger \textbf{Jaco hand}~\cite{JACOArm}. This already challenges cross-embodiment visual tracking as there is already a significant embodiment gap.
The following 3 domains are all in the real world. In our \textbf{real robot} environment, which is similar to our canonical simulation, we have a Sawyer arm with a Robotiq 2-finger gripper mounted behind a basket with 3 objects. These are the 3D printed versions of the procedurally-generated objects we use in simulation.
The gripper is fitted with custom 3D printed fingertips. Similar to the simulation, there are two cameras at the front left and right of the basket which provide $128\times128$ resolution images to the agents. However, the low-level appearance and the physics are quite different
to those in our simulated environment. 
In our most challenging environments, we use the same real environment but we ourselves instead directly manipulate the objects with a \textbf{pick-up stick} or our \textbf{hand}.

\subsection{Dataset for MIR Training} 
\label{sec:dataset}
As explained in section~\ref{sec:CEVI}, learning manipulator-independent representations with our method requires temporally-aligned sequences. In our work, this is done by utilizing simulation and domain randomization to create different visual versions of the same trajectories that can be used for our perceptual training with TSCN and CD-GCP.

In total we generated $7,194$ canonical trajectories of reaching, grasping, lifting, and stacking with all combinations of the three objects in the basket. These were replicated in the `invisible arm', `domain-randomized', and `arm-randomized' environments. The canonical trajectories were discarded and, as illustrated in Figure~\ref{fig:methods}, the trajectories from the `domain-randomized' environment were paired with the equivalent trajectories of each of the other domains for a total of $14,388$ episodes. Out of these, $10,792$ paired trajectories were used for MIR training. The rest were held out for model selection and hyperparameter tuning such as picking the learning rate, batch size and number of epochs. 
The episodes in our dataset were collected by multiple stacking policies trained with a staged reward for reaching, grasping, lifting, and stacking. They were trained in the canonical simulation environment from fully-observed states, separately for each color pair. As we used policies at multiple points during their training, we have a training dataset that consists of a diverse set of manipulation trajectories, and we assume that most of the state space of object configurations is reasonable represented.  A sample trajectory is $200$ time-steps, i.e.\ 10 seconds long with a control frequency of 20Hz. The policies were trained using  Maximum a Posteriori Policy Optimization (MPO)~\cite{abdolmaleki2018maximum} following the implementation details of \citet{jeong2019self} which solved a similar stacking task.

\subsection{Evaluation Trajectories}
In order to evaluate the cross-embodiment aspect of imitation in a principled way we picked a single type of behavior for the trajectories to imitate. 
We chose that to be stacking one geometric object on top of another in the presence of a third distractor object. There are multiple reasons for this decision. 
Firstly, having the same type of demonstrations makes it possible to compare the complexity of different environments and how our method deals with increasing levels of domain gap.
Secondly, instead of analyzing our results with respect to a tracking reward that is hard to interpret, a multi-stage behavior allows us to intuitively illustrate our analysis of imitation success in terms of achieving the different progressive stages of the demonstrated behaviors. In the Tables~\ref{tab:mir_comparison} and~\ref{tab:ablation}, these stages are `\textbf{lifting} the top object' and `\textbf{stacking} the top object onto the bottom one.' We excluded `reaching the top object' for clarity, as it is a simple motion that does not involve manipulation.
Thirdly, we believe stacking is a task that would challenge existing trajectory tracking methods even in the same-embodiment setting, provided it is performed in an environment with realistic physics and complex control. The action space of our agents is the velocities of the robot joints and gripper, and the physics of our simulated environment are reasonably close to the ones in the real world. Stacking involves multi-object interaction and rich contact dynamics, especially with the geometric objects we have chosen. It clearly exposes the difficulty of imitation once we move away from simple, almost planar tasks typically showcased in works on imitation from observation only. Finally, stacking allows for relatively simple automated evaluations in the real world.
Note that choosing a single behavior for our evaluation setting is without loss of generality: the imitating agent is behavior-agnostic and is always just tasked with following a given trajectory with high precision. Behavior generality is only limited by the type of perceptual training and the data used for it. In Appendix~\ref{app:qualitative} and in our supplementary material we provide examples of how our method is able to generalize beyond the data it was trained with.

We chose to collect $10$ sequences from each of the five demonstration domains used during evaluation: `invisible arm', `Jaco hand', 'real robot', 'pick-up stick', and 'human hand'. Those were chosen to be of an increasing domain gap with the canonical simulation environment of our imitating agent, in terms of embodiment, vision and dynamics.  For the `invisible arm' domain, which was also used during MIR training, we used held-out trajectories collected with the stacking RL policies as discussed above. The rest of the domains were unseen by the visual feature encoders used for imitation. For most of the held-out demonstration environments (`Jaco hand', `pick-up stick', and `human hand'), the trajectories were manually collected via teleoperation or by hand. For the `real robot' we instead used successful trajectories from the zero-shot evaluation of a variant of the policy we used for MIR training, trained in the domain-randomized environment.

\section{Experimental Evaluation}
\label{sec:results}
In this section, we demonstrate the efficacy of our method, and provide comprehensive quantitative analysis on our design choices and comparison with a few strong baselines. As discussed above, we evaluate how well each perceptual method aids RL-based imitation of 10 trajectories from each of the five demonstration environments. During imitation, in order to start from a similar initial condition, we manually set the positions of objects similar to the beginning of the target trajectory. Furthermore, our simulation environment exhibits non-deterministic behavior when resetting to a fixed initial condition. Such stochasticity is introduced due to our modelling of the robot. Therefore, for each trajectory collected, we evaluate each agent $100$ times and report the lifting success rate and stacking success rate for a total of $1,000$ trajectories for each domain-method combination. 
As mentioned above, lifting and stacking success are reported to provide a clear picture of what we would like to test for during trajectory following, i.e.\ that the environmental changes are the same in the demonstration and the trajectory produced by an agent.
First we compare our method with several strong baselines for cross-embodiment trajectory tracking, and then we'll discuss the contribution of each component in MIR.

\begin{table*}[h]
\centering
    \begin{tabular}{l||r|r|r|r|r||r|r|r|r|r}
    \toprule
    & \multicolumn{5}{c||}{\textbf{Lifting Success}}
    & \multicolumn{5}{c}{\textbf{Stacking Success}}  \\
        &    \textbf{invisible} &   \textbf{jaco}&\textbf{robot} & \textbf{stick}& \textbf{hand}
        &    \textbf{invisible} &   \textbf{jaco}&\textbf{robot} & \textbf{stick}& \textbf{hand} \\
    \textbf{Method}       &        &        &      &    &    &    &    &    &   \\
    \midrule
    \quad TCN \cite{sermanet2018time}         
                    &   79\%            &  85\%           & 27\%                  & 12\%                       & 0\%  
                    &   0\%             &  0\%            & 0\%                   & 0\%                       & 0\%  \\    
    \quad TSCN        
                    &   \textbf{100\%}  &  \textbf{100\%}   &\textbf{100\%}         & 20\%                      & 40\%
                    &   0\%             &  67\%             &0\%                    & 0\%                       & 7\% \\
    \quad CD-GCP        
                    &   86\%            &  44\%             &50\%                   & 20\%                       & 0\%
                    &   0\%             &  8\%              &0\%                    & 0\%                       & 0\% \\

    \quad CD-GCP + TCN  
                    &   90\%            &  \textbf{100\%}   &89\%                   & 22\%                      & 20\%
                    &   20\%            &  38\%             & 8\%                   & 0\%.                      & 8\%  \\
    
    \quad MIR (CD-GCP + TSCN)
                    & \textbf{100\%}    &  \textbf{100\%}   & \textbf{100\%}        & \textbf{44\%}             & \textbf{50\%}
                    & \textbf{38\%}     &  \textbf{81\%}    & \textbf{29\%}                  & \textbf{17\%}             & \textbf{11\%} \\    
    \bottomrule
    \end{tabular}

\caption{Ablation study of each component in our method. Evaluation is performed on 100 attempts of following for each of the 10 demonstrated stacking trajectories on five domains: invisible-arm, jaco hand, real robot, pick-up stick and human hand. The metrics are the success rate of imitating lifting and stacking behaviours.}
\label{tab:ablation}
\vspace{-0.8cm}
\end{table*}
\subsection{Comparison of Imitation Performance}
We perform cross-embodiment imitation using a few strong baseline representations and our MIR method using an MPO agent as described in Section~\ref{sec:imitation}. We report the performance of successful imitation of trajectories at two stages, lifting and stacking success, in Table~\ref{tab:mir_comparison}. 

The first group of baselines doesn't utilize the paired nature of the trajectories across domains. These are: na\"{i}ve  Goal-Conditioned Policies ({GCP}), as described in section~\ref{sec:MIR}; and Temporal Distance Classification (TDC)~\cite{aytar2018playing}, which learns a representation by classifying temporal distance between any given pair of observations within an episode. For a more detailed summary of these baselines see Appendix~\ref{app:baselines}. Note that TDC uses all three training domains but not the temporally paired nature of the data. GCP is trained using target goals within the same domain, and used the `domain-randomized' and `arm-randomized' data, as the `invisible arm' data in a single-domain setting would hurt its performance given that GCP is tasked with predicting actions given, among others, the position of the gripper. 
The second group of baselines is domain alignment methods which explicitly utilize paired trajectories. These are: Time-Contrastive Networks (TCN)~\cite{sermanet2018time}, as described in section~\ref{sec:MIR}; and Cross-Modal Distance Classification (CMC)\cite{aytar2018playing}, which learns a representation by classifying temporal distance between any given pair of observations across domains (see Appendix \ref{app:baselines}). Our MIR method also belongs in this category.
Overall, both for imitating lifting and stacking behaviors, domain alignment methods perform much better. MIR clearly achieves the best performance in all test domains for both lifting and stacking stages.

MIR perfectly imitates simulated  `Jaco hand' and `invisible arm' sequences for lifting, and achieves $81\%$ success rate for `Jaco hand' and $38\%$ performance for `invisible arm' for completing the full trajectory of stacking. Note that `Jaco hand' is never seen during training of MIR. Although `invisible arm' is used during MIR training, following demonstrations in this environment remains challenging: the manipulator is invisible, and the camera observation of the scene remains static while there is no contact with the objects. Because of that, the first goals are likely to be identical to the first frame and not informative. As imitation relies solely on vision, this makes it harder for the arm to reach the object that needs to move later in the demonstration.

As the domain gap increases in both visual perception and dynamics the performance of methods decrease across the board for real-world sequences. This clearly demonstrates how hard it is to generalize to unseen real-world settings. In addition, the increasing complexity of the gripper morphology (i.e. real robot, pick-up stick, and human hand) also affects the performance in a negative way for all methods. However, MIR still achieves excellent imitation performance for the lifting stage of real sequences. More importantly, MIR is the only method that can successfully imitate some of the real-world sequences for the full stacking trajectories in all three real-world settings. 

Imitating a stacking behavior (i.e.\ a long horizon episode with multi-object contact-rich dynamics) proves to be significantly harder than lifting, and only CMC and MIR achieve considerable performance when it comes to stacking. This clearly demonstrates the need for further research in visual imitation of multi-stage contact-rich sequences.

\subsection{Ablation Study}
As shown in Table~\ref{tab:mir_comparison} our proposed MIR method is able to achieve better imitation compared to other baselines we investigated. As MIR is a combination of TSCN and CD-GCP training, this begs the following question: are all the components of MIR training crucial?

We investigated whether certain aspects of our proposed MIR are indeed important and present our findings in Table~\ref{tab:ablation}. We first examined whether combining TSCN and CD-GCP for our proposed MIR method is indeed obtaining better representations for imitation, compared to the two components individually. We also looked into the importance of the temporal smoothness aspect of TSCN. We did so by replacing our TSCN loss with the standard TCN~\cite{sermanet2018time} one, since former is a temporally-smooth variation of the latter, as discussed in Section~\ref{sec:MIR}. It is clear from the results that using TSCN is preferable to TCN, both when using it on its own and when combining it with CD-GCP, with performance gains that are quite significant and consistent across the domains and stages of imitation.

In this experimental section we have demonstrated that our MIR method has the right characteristics for cross-embodiment visual imitation with RL. The interested reader can find a t-SNE visualization of the learned embedding spaces and more qualitative examples in the Appendix and our anonymized project website\footnote{\url{https://sites.google.com/view/mir4vi}}. %

\section{Conclusion}
\label{sec:conclusion}

In this work we explored cross-embodiment visual imitation of robotic manipulation trajectories.
We demonstrated the importance of the representation space in visual imitation, and introduced manipulation-independent representations (MIR) as a suitable candidate that can support successful imitation of behavior demonstrated by previously unseen manipulator morphologies. This included imitating a real human hand with a simulated robot. Through MIR, we emphasised what is important in learning manipulator and task independent representations for imitation. They need to be robust to domain gaps, and mainly focus on the change of object configurations, while maintaining the ability to capture the rough notion of a manipulator. They also need to be temporally smooth and actionable. These were addressed by introducing TSCN for the former and CD-GCPs for the latter. All these claims were validated through extensive quantitative experimentation.

\section*{Acknowledgments}

We gratefully acknowledge our team member Rae Chan Jeong for helping with simulator modification, David Khosid, Claudio Fantacci, and Federico Casarini for their support in manipulating real robots and collecting human demonstrations and Serkan Cabi for fruitful discussions and proofreading.

\bibliographystyle{plainnat}
\bibliography{references}

\clearpage
\appendix

\subsection{t-SNE Visualization}
\begin{figure}[t]
\centering
\includegraphics[width=0.45\textwidth]{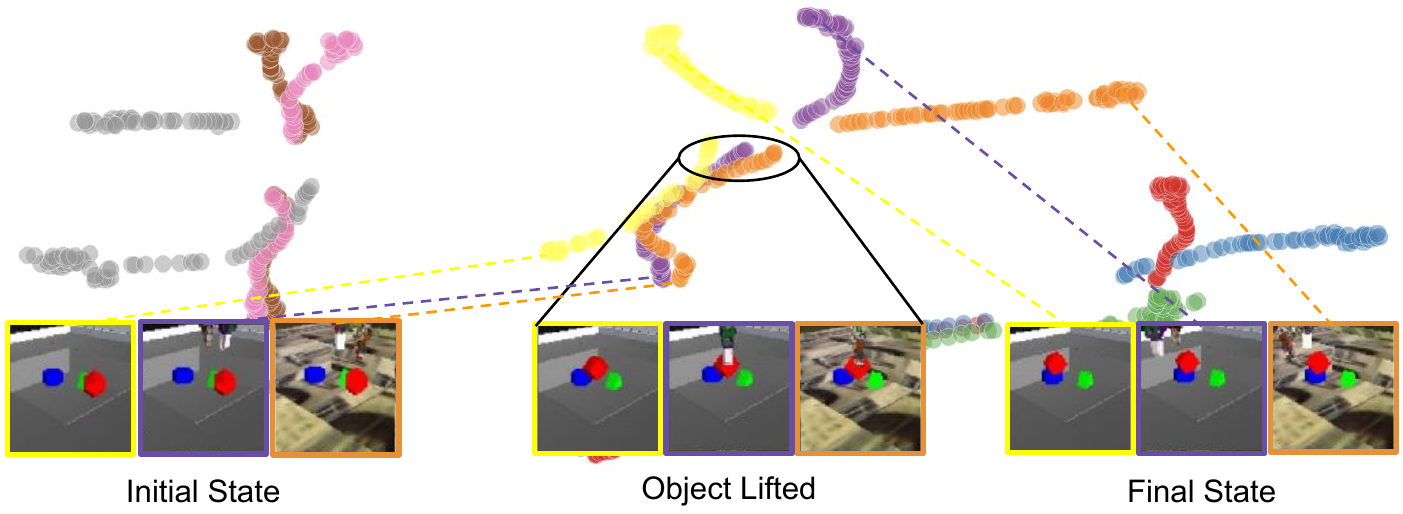}
\caption{t-SNE Projection of 3 triplets of trajectories, where each triplet consists of paired trajectories rendered in 3 visually different domains: with an `invisible arm', with full domain randomization and with domain randomization only for the robotic arm. The features used were learned with cross-domain goal-conditioned policy (CD-GCP) combined with TSCN. The 3 triplets are clearly separated from each other and the trajectories from the different domains seem to align well, especially when there is an environment change that doesn't include the arm, a desired feature for our representations. See text for more details.}
\label{fig:exp_tsne}
\end{figure}

In order to gain better insight into embeddings we learn with the MIR methods, we plotted the t-SNE projection for each method on three aligned triplets, for a total of 9 trajectories from the three different domains used for training our manipulator-independent representations (MIR): domain-randomized, arm-randomized, and `invisible arm'. Figure~\ref{fig:exp_tsne} depicts the alignment achieved by learning the representations of MIR. As one can see in the figure, where we depict the frames at certain points in the trajectories for a given triplet, the trajectories are aligning fairly well even though they are from different domains, while staying further from the other two trajectory-triplets. Alignment seems particularly successfully at the point of an environment change, i.e. when the red object is being lifted. 

We also provided video t-SNE visualizations in the video provided in supplementary materials.

\subsection{Qualitative Evaluation Videos}
\label{app:qualitative}
We provide example imitation trajectories that are part of our quantitative evaluation. All the experiments are using the pretrained MIR as described in the main paper. Our method successfully learns to imitate each trajectory despite being out of distribution. A brief description of the qualitative evaluations are provided below:

\textbf{Example trajectory following for each domains.} We firstly show examples of third-person demonstrated trajectories and how our agents managed to follow them. A few examples are presented for each of the testing domains, namely `invisible arm' simulation, `Jaco hand' simulation, shadow hand simulation and real-world robot.

\textbf{Stack and unstack.} Stacking red object on top of blue object, releasing the blue object and rising up, grasping the blue object again and returning it to its original location.

\textbf{Stack with obstacles.} A fixed obstacle is added into the basket as a distractor, while the obstacle does not exist in the demonstrated trajectory. Note that obstacles are not seen during MIR training.

\begin{figure}[t]
\centering
\includegraphics[width=0.45\textwidth]{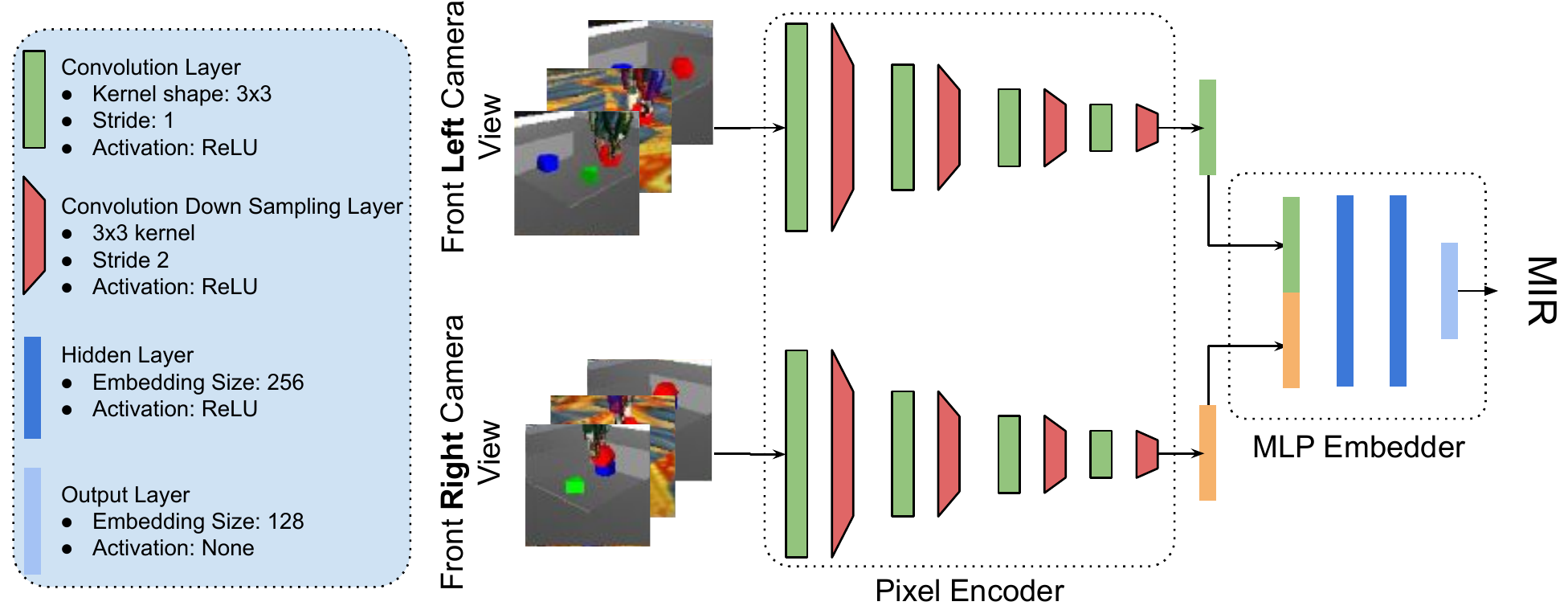}
\caption{MIR Encoder Structure. Our agents can access both front cameras in the environment. For each camera observation, a convolutional encoder is applied to extract a 128-dimension feature. A 3-layer MLP is then applied to the concatenated features and output a 128-dimension representation for the final MIR. Configurations of different layers are shown one the left.}
\label{fig:encoder}
\end{figure}

\begin{figure*}[]
    \centering
    \includegraphics[width=\textwidth]{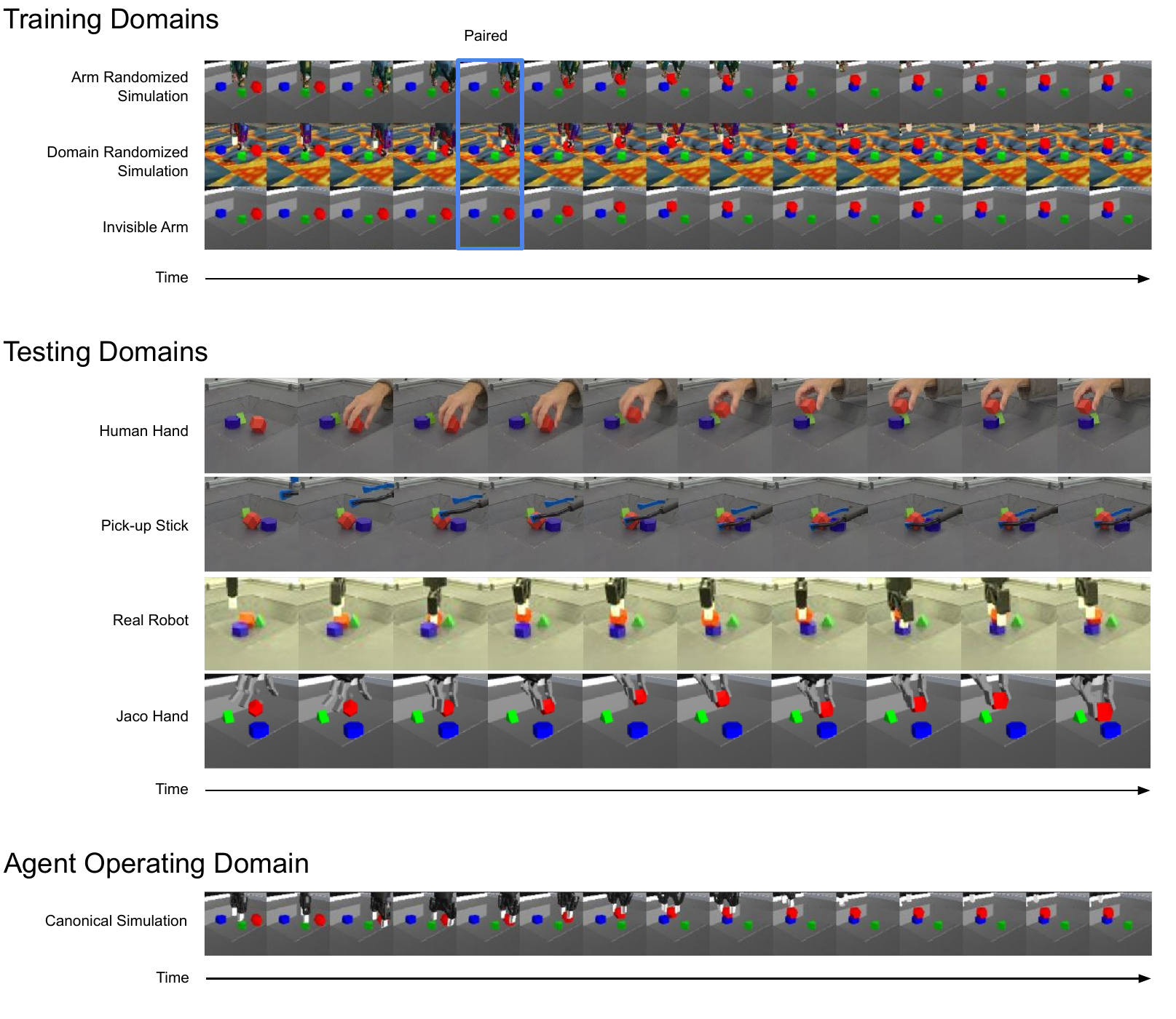}
    \caption{Filmstrips of MIR training and testing domains. \textbf{In training domains}, examples of paired trajectories used for manipulator independent representation learning are shown. A sub-sampled filmstrip with 10 steps interval is presented from left to right. From top to bottom, the environment involved are arm randomized simulation, domain randomized simulation and simulation with invisible hand. \textbf{In testing domains}, examples of demonstrated trajectories are presented. Similarly, sub-sampled filmstrips are shown for human hand, pick-up stick, real robot, and Jaco hand domains. Note that there is no correlation between any trajectories in testing domains. The canonical environment, where our simulated robot agent is operating on, is also shown at the bottom.}
    \label{fig:three-traj}
\end{figure*}

\subsection{Baselines}
\label{app:baselines}
As TCN is already covered in the main text, here we'll describe the other two baselines compared to our method.

{\bf TDC~\citep{aytar2018playing}}. As it is evident from its name, temporal distance classification (TDC) predicts the temporal distance between any given observation pair within the same sequence. 
TDC sets the distance prediction as a classification problem where the classes are different distance intervals. We follow the same setting described in \citep{aytar2018playing} and set the distance interval categories as $\{[1], [2], [3-4], [5-20], [21-200]\}$, which suits well to our setting as we operate over episodes of $200$ time-steps. 
To define the loss, given the observation pair $(x_i,x_{i+d})$ we concatenate the features and run them through a two-level MLP to predict the distance interval. We use the distance $d$ to set the true label. The optimized loss function is a softmax cross-entropy for this multi-class classification problem.

{\bf CMC\citep{aytar2018playing}}. Cross-modal distance classification (CMC) predicts the temporal distance between any given cross-modal (i.e. cross-domain) observation pair. \cite{aytar2018playing} uses this method to learn representations by aligning audio and images in game videos. In our setting modalities are the pairs of randomized domains. For instance in the cross-domain pair $(x_i,\bar{x}_{i+2})$ the temporal distance would be $2$. CMC also sets the distance prediction as a classification problem where $[0]$ distance is also included in the distance interval categories used for TDC. To define the loss, given the cross-domain pair $(x_i,\bar{x}_{i+d})$ we concatenate the features and run them through a two-level MLP to predict the distance interval. We use the distance $d$ to set the true label. The optimized loss function is a softmax cross-entropy for this multi-class classification problem.

\subsection{Network Structure for MIR}

In our environment, an agent can access two front cameras. The two cameras are beneficial to help disambiguate 3D positions of the robot arm and the objects. We extract image representations separately from each camera using the same convolutional encoder. As shown in Figure \ref{fig:encoder}, the encoder has a 4-step repetition of one 3x3 convolutions each follow by a down-sampling 3x3 convolution layer with stride 2. A rectified linear unit (ReLU) is applied after each convolution layer. At each down-sampling step we double the number of feature channels, where first step has channel size 8. The output of a pixel encoder is a 128-dimension feature which is obtained through a final linear layer. Features of both camera views are concatenated channel-wise followed by a 3-layer multilayer perceptron (MLP) network. Each hidden layer is the MLP has 256 channels and the output of the MLP is a 128-channel embedding. The configuration of layers are shown in Figure~\ref{fig:encoder}. For GCP and CD-GCP training we also use a separate goal encoder which shares the same structure described above but weights are not shared. Additionally a 3 layer MLP with $[256, 256, 5]$ channels is used as the policy head to predict the actions.

\end{document}